\documentclass[a4paper,fleqn]{cas-dc}

\usepackage[authoryear,longnamesfirst]{natbib}
\usepackage{amsmath}
\usepackage{algorithm}
\usepackage{algpseudocode}
\usepackage{siunitx}

\def\tsc#1{\csdef{#1}{\textsc{\lowercase{#1}}\xspace}}
\tsc{WGM}
\tsc{QE}
\tsc{EP}
\tsc{PMS}
\tsc{BEC}
\tsc{DE}
\begin{document}
\let\WriteBookmarks\relax
\def\floatpagepagefraction{1}
\def\textpagefraction{.001}

\shorttitle{Dynamic Spatio-Temporal  Summarization using Information Based
Fusion}

\shortauthors{Humayra Tasnim et~al.}

%\title [mode = title]{Feature-Based Temporal Data Summaries using Information Based Fusion} 
\title [mode = title]{Dynamic Spatio-Temporal Summarization using Information Based Fusion} 
%%\tnotemark[1,2]

%%\tnotetext[1]{This document is the results of the research
  %% project funded by the National Science Foundation.}

%%\tnotetext[2]{The second title footnote which is a longer text matter
  %% to fill through the whole text width and overflow into
   %%another line in the footnotes area of the first page.}

\author[1]{Humayra Tasnim}[type=editor,
                        orcid=0000-0002-7796-7717]

\cormark[1]


\ead{htasnim30@unm.edu}

\ead[url]{htasnim.github.io}

\credit{Conceptualization, Methodology, Software, Data curation, Validation, Visualization, Writing - Original draft preparation, Writing - review and editing }

\affiliation[1]{organization={University of New Mexico},
    %addressline={Radarweg 29}, 
    city={Albuquerque},
    % citysep={}, % Uncomment if no comma needed between city and postcode
    state={New Mexico},
    postcode={87131}, 
    country={United States}}

\author[2]{Soumya Dutta}[type=editor,orcid=0000-0001-5030-9979]

% Principal Corresponding author indication
%\cormark[2]

% Footnote of the second author

\ead{soumyad@cse.iitk.ac.in}

\ead[url]{soumyadutta-cse.github.io}

\credit{Conceptualization, Supervision, Methodology, Validation, Writing - Original draft preparation, Writing - review and editing}

\affiliation[2]{organization={Indian Institute of Technology Kanpur},
    %addressline={G66M+W5J, Kalyanpur}, 
    %city={Kanpur},
    % citysep={}, % Uncomment if no comma needed between city and postcode
    state={Uttar Pradesh},
    postcode={208016}, 
    country={India}}

\author[1]{Melanie Moses}[type=editor]
\ead{melaniem@cs.unm.edu}
\ead[URL]{moseslab.cs.unm.edu}

\credit{Supervision, Writing - review and editing}

%\affiliation[3]{organization={Sayahna Foundation},
 %   city={Jagathy},
    % citysep={}, % Uncomment if no comma needed between city and postcode
  %  postcode={695014}, 
   % state={Trivandrum},
    %country={India}}

%\affiliation[3]{organization={STM Document Engineering Pvt Ltd.},
 %   addressline={Mepukada}, 
  %  city={Malayinkil},
   % % citysep={}, % Uncomment if no comma needed between city and postcode
    %postcode={695571}, 
    %state={Trivandrum},
    %country={India}}

\cortext[cor1]{Corresponding author}
%\cortext[cor2]{Principal corresponding author}

\iffalse
\fntext[fn1]{This is the first author footnote. but is common to third
  author as well.}
\fntext[fn2]{Another author footnote, this is a very long footnote and
  it should be a really long footnote. But this footnote is not yet
  sufficiently long enough to make two lines of footnote text.}

\nonumnote{This note has no numbers. In this work we demonstrate $a_b$
  the formation Y\_1 of a new type of polariton on the interface
  between a cuprous oxide slab and a polystyrene micro-sphere placed
  on the slab.
  }
\fi
\begin{abstract}
In the era of burgeoning data generation, managing and storing large-scale time-varying datasets poses significant challenges. With the rise of supercomputing capabilities, the volume of data produced has soared, intensifying storage and I/O overheads. To address this issue, we propose a dynamic spatio-temporal data summarization technique that identifies informative features in key timesteps and fuses less informative ones. This approach minimizes storage requirements while preserving data dynamics. Unlike existing methods, our method retains both raw and summarized timesteps, ensuring a comprehensive view of information changes over time. We utilize information-theoretic measures to guide the fusion process, resulting in a visual representation that captures essential data patterns. We demonstrate the versatility of our technique across diverse datasets, encompassing particle-based flow simulations, security and surveillance applications, and biological cell interactions within the immune system. Our research significantly contributes to the realm of data management, introducing enhanced efficiency and deeper insights across diverse multidisciplinary domains. We provide a streamlined approach for handling massive datasets that can be applied to in situ analysis as well as post hoc analysis. This not only addresses the escalating challenges of data storage and I/O overheads but also unlocks the potential for informed decision-making. Our method empowers researchers and experts to explore essential temporal dynamics while minimizing storage requirements, thereby fostering a more effective and intuitive understanding of complex data behaviors.
\end{abstract}

% \includegraphics{figs/grabs.pdf}

\begin{highlights}
\item The research introduces a novel and adaptable method for interpreting informative features of large scale spatiotemporal data, applicable to diverse datasets from different domains.  
\item The proposed technique identifies key informative timesteps and uses information-based fusion to summarize salient patterns of important features in less informative timesteps, leading to significant storage optimization without compromising data insights. 
\item Proposed approach facilitates effective visualization, tracking, and analysis of temporal changes in the datasets.
\end{highlights}

% Keywords

\maketitle

\section{Introduction}

In today's data-driven world, the exponential growth in data generation has brought forth significant challenges for storage and associated I/O overheads. Modern supercomputing capabilities have reduced the computational cost of producing data, resulting in the creation of massive datasets at an accelerating pace [\cite{reed2015exascale,childs2015data}]. Many of these datasets exhibit a dynamic temporal nature, consisting of multiple timesteps that may be redundant or offer minimal new information when analyzed timestep by timestep. The storage burden of retaining these less informative timesteps poses a critical issue, calling for novel techniques that can efficiently manage such large-scale time-varying datasets.

To address this challenge, we propose a data summarization technique that aims to minimize the storage overhead while preserving the vital temporal dynamics of the dataset. We also emphasize visualizing these dynamics by tracking changes over time. Our approach involves an dynamic spatio-temporal technique, which identifies both key timesteps and redundant timesteps. We store the key timesteps and summarize the redundant multiple timesteps into one, highlighting the salient features. The summarization technique effectively ensures storage reduction with minimal loss. We propose to use information-theoretic measures namely the Specific Mutual Information or SMI-guided fusion for the summaries. 

The core idea of the summarization technique is to identify informative temporal features within the redundant timesteps and fuse them using principles from information theory. By selecting the most relevant features from the redundant timesteps and summarizing them through information-guided fusion, we ensure that the temporal dynamics are retained. This approach not only optimizes storage requirements but also facilitates the visualization and tracking of information change over time, providing valuable insights into the underlying data patterns.

%In this work, our research aim is to develop a method that efficiently handles large-scale time-varying datasets across various domains. By leveraging the principles of the information theory fusion process, we can reduce storage overhead without compromising the dataset's essential features. The proposed approach has broad applications across multidisciplinary fields, from simulations in chemical looping reactors to analyzing dynamic interactions between immune cells in lymph nodes and tracking incidents in surveillance footage.

In the paper, we present the details of our data summarization technique, dynamic spatio-temporal framework, and the information-guided fusion process. We demonstrate the technique's versatility by applying it to different datasets, including scalar data from particle-based simulations and image data from surveillance footage and cellular interactions. Our results show significant storage reduction without compromising critical insights, proving the effectiveness of our approach for efficient data management and visualization.

Our research aim in this work is to develop a method that efficiently handles large-scale time-varying datasets across various domains. Our solution seeks to bridge the data reduction landscape by providing approaches to effectively manage large and intricate temporal datasets. Our goal is to achieve this while retaining essential features and enabling robust visualization techniques. By leveraging the power of information theory, we aim to contribute to a more efficient and insightful data management strategy, especially in the context of dynamic and multifaceted datasets.

The contributions of the paper are:
\begin{itemize}
    \item  Develop a dynamic spatio-temporal summarization (DSTS) technique for large-scale time-varying datasets.  The summary provides three features: key timesteps, fused timesteps, and holistic visual representation of information change. 
    \item Propose and demonstrate that the specific mutual information (SMI) measure "Surprise" provides the most promising outcomes for the fusion technique.
    \item Demonstrate the versatility and effectiveness of the proposed technique (acm) through application to diverse datasets, including chemical simulations, surveillance footage, and cell interactions in the immune system.
    \item Explore the impact of the technique in optimizing data storage with minimal data loss. % and visualization highlighting the spatio-temporal dynamics of the analyzed datasets.
    %\item \mm{the surprise method is the most effective one. comparison with other information theory method. Replace this with 2}
\end{itemize}

The paper's structure is as follows: Section 2 reviews related works on data summarization and information theory-based approaches. Section 3 outlines the methodology for the framework %, encompassing the dynamic spatio-temporal framework and the information-guided fusion process.
Section 4 demonstrates the results of implementing the technique across multi-domain datasets. Section 5 delves into the implications of our findings and the research's importance in diverse application domains. Lastly, Section 6 concludes the paper by highlighting possible future research avenues.

\label{intro}
\section{Related Works}

Efficiently managing and analyzing large-scale time-varying datasets has attracted considerable attention across multiple disciplines, leading to various strategies to address the challenges posed by data growth. In this work, we focus on identifying and storing informative key timesteps while summarizing less informative ones by fusion. The summaries reduce storage overhead while the key and fused timesteps preserve data dynamics. Various data compression and reduction techniques have been explored in the context of large-scale data analysis, aligning with the goals of the proposed research. There are techniques that represent data reduction where reduced data is stored in place of the raw data. Cinema [\cite{Ahrens2014}] is an in situ data reduction and visualization approach that utilizes image-based representations to capture essential features of time-varying datasets. Techniques like lossless and lossy compression methods are commonly used for data reduction.  References [\cite{vidhya2016review,cappello2019use,di2016fast}] present various compression algorithms and their application in handling large datasets. Other reduction technique includes sub-sampling [\cite{woodring2011situ,wei2018information}] and distribution-based summaries [\cite{dutta2016situ, ye2016situ}]. The abovementioned methods are widely recognized for their data reduction and compression effectiveness. However, these approaches solely store the reduced timesteps. In our work, we need to retain both the raw and reduced timesteps to comprehensively capture information changes over time and preserve essential data dynamics.

Identifying key time steps is essential in analyzing time-varying data and advancing scientific visualization. There are numerous approaches [\cite{zhou2018key,myers2016partitioning,tong2012salient, hu2011survey}] that have been proposed for identifying key time steps in large time-varying datasets. %including videos and sequential imaging.
These studies focus on only capturing key time steps without the summarization capacity. Some studies focused on data reduction such as: in [\cite{akiba2006simultaneous}]  similar timesteps are grouped and one is selected, in \cite{wang2008importance} salient timesteps were selected by comparing dissimilarity with previous timestep. These studies did not include any summarization techniques which we incorporated in our proposed work along with selecting key time steps.

The development of fusion techniques [\cite{castanedo2013review}] for large-scale spatio-temporal datasets has been a popular field of interest for researchers across various domains like remote sensing [\cite{li2020spatio,chen2015comparison,nguyen2014spatio}], geoscience [\cite{wu2021spatially,ma2020spatio,tang2020quantifying}], network architectures [\cite{kashinath2021review, shah2017spatiotemporal}], and computer vision[\cite{duzceker2021deepvideomvs,wang2020virtual,xue2017bayesian}].In the realm of computer vision, various strategies for data summarization using fusion techniques have been explored, including Gaussian entropy fusion [\cite{fu2010multi}] and probabilistic skimlets fusion [\cite{zhang2014probabilistic}]. Additionally, deep learning architectures have been applied for summarization tasks [\cite{zhong2019video}]. Notably, the proposed work takes a unique approach by leveraging information-theoretic fusion for data summarization. Unlike some existing techniques, our approach doesn't need training and can be readily applied to large-scale datasets. Moreover, it requires minimal computational costs, rendering it effective for both in situ and post hoc analysis. The feasibility of in situ application is showcased in our preliminary research [\cite{Dutta2021}].

Information theory [\cite{Cover2006, 720531}] concepts such as entropy and mutual information [\cite{Shannon1948,gray2011entropy}]  have been employed to measure the relationships between variables in data across multiple computational domains [\cite{Moore2018,Pilkiewicz2020, sbert2022information, Tasnim2018}]. Mutual information is extensively applied in the field of feature selection, exploration, extraction, and tracking [\cite{Biswas2013, Dutta2016,Tasnim2022}]. Image registration is another popular %area of interest for information theory 
application [\cite{maes1997multimodality,viola1997alignment,hill2001medical,Cahill2010}]. Mutual information, as well as its decomposition measures like specific mutual information and pointwise mutual information, have been widely used in multimodal data fusion [\cite{Bramon2012}], graphics, and visualization [\cite{wang2008importance,dutta2016situ,Dutta2017,Dutta2017a,Isola2014,Bramon2013a,Akiba2007,chen2016information}]. Other use cases for information theory include view selection [\cite{viola2006importance}], feature similarity [\cite{bruckner2010isosurface}], transfer function and design [\cite{Ruiz2011, Bramon2013, Haidacher2008}], and data sampling [\cite{Dutta2019,wei2018information}]. 

The works we discussed here collectively contribute to the proposed approach's design, development, and motivation. Our research draws inspiration from these methodologies and integrates information-theoretic principles to develop an innovative solution. %This solution seeks to bridge the data reduction landscape by providing approaches to effectively manage large and intricate temporal datasets. Our goal is to achieve this while retaining essential features and enabling robust visualization techniques. By leveraging the power of information theory, we aim to contribute to a more efficient and insightful data management strategy, especially in the context of dynamic and multifaceted datasets.
 \label{related}
\section{Information-Driven Framework for Feature-Based Temporal Data Summaries} 

\begin{figure*}
	\centering
		\includegraphics[scale=.40]{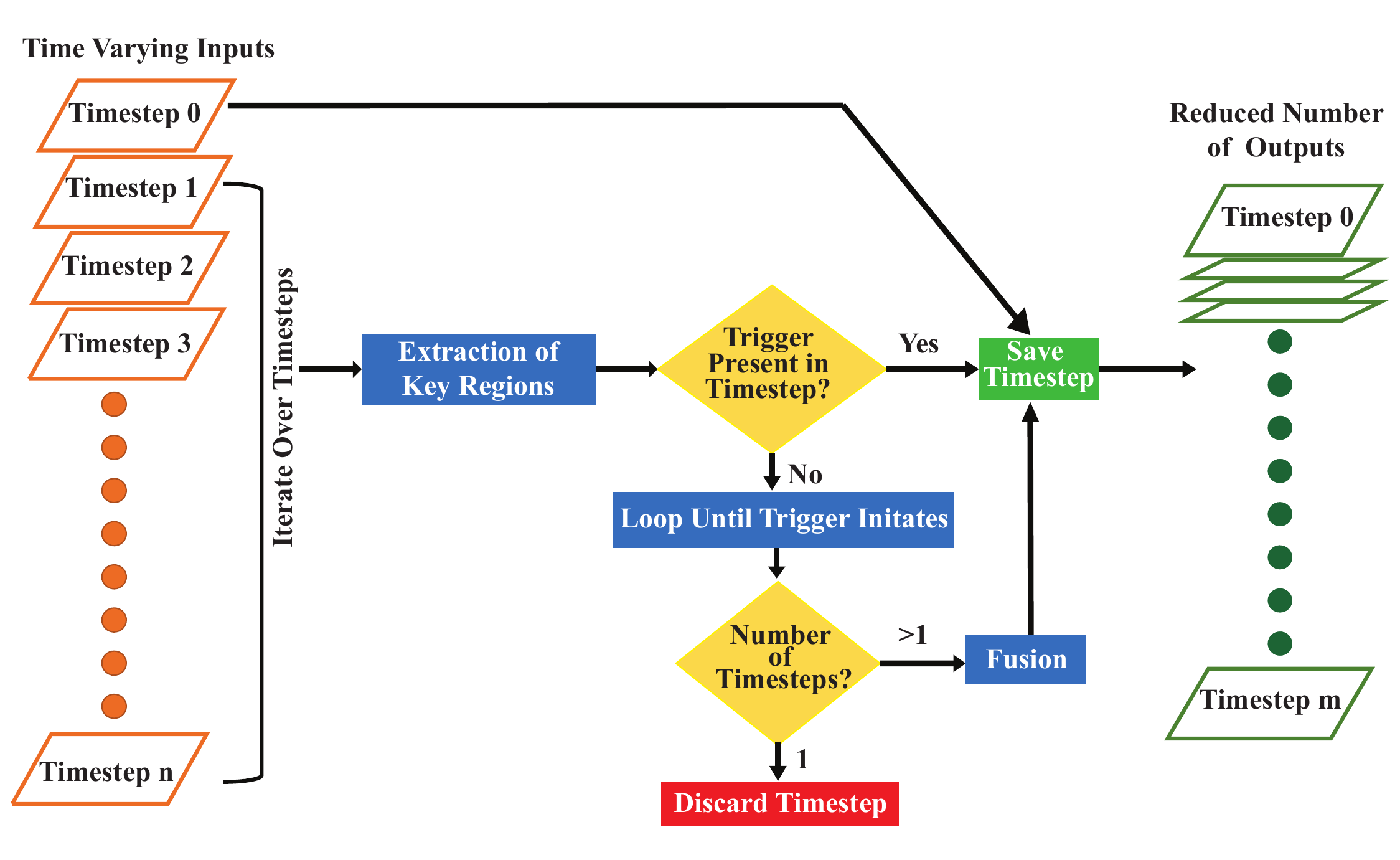}
	\caption{Schematic diagram of the workflow. We have used the standard computational flowchart [\cite{sipser13}] symbols for representations: input/output, process, decision, and arrows indicating relationships between symbols.}
	\label{fig:workflow}
\end{figure*}

\subsection{Framework Workflow}
\begin{figure*}
\centering
\includegraphics[scale=.19]{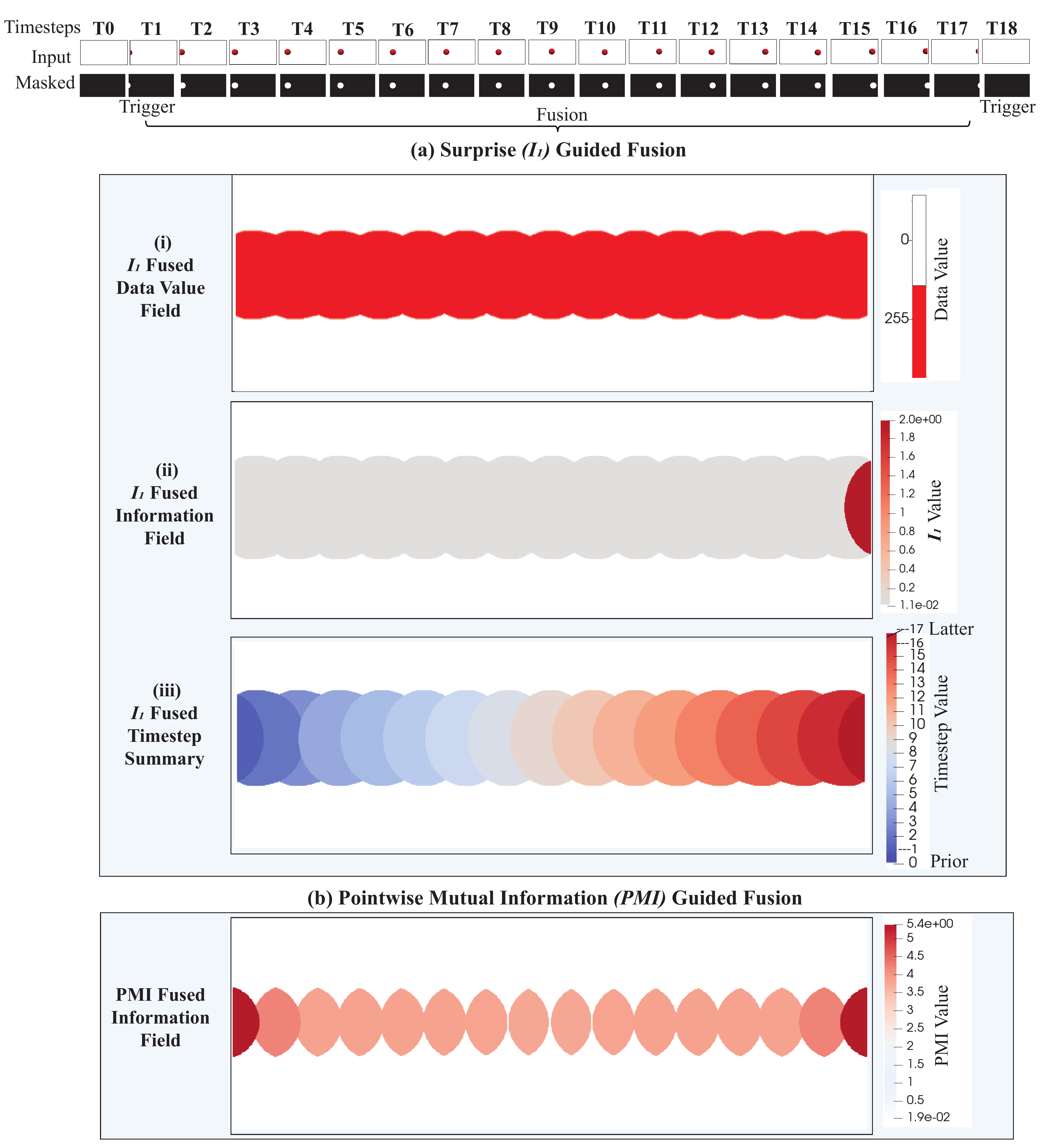}
\caption{Illustration of a simulated rolling ball with 19 timesteps (T0 -T18). At each timestep, the ball moves 0.5 units to the right. Timesteps are 2D RGB images with 800 x 400 dimensions and samples ranging from 0 to 255 (input row). The masked row presents binary images with values 0 (no ball) and 255 (ball) after ball segmentation. Timesteps T1 to T17 are fused using (a) Surprise ($I_1$) guided fusion and (b) PMI guided fusion. a(i) shows the $I_1$ fused data value field of the fused timesteps(0 white and 255 red). a(ii) shows $I_1$ fused information value field.  a(iii) displays $I_1$ fused timestep summary with numbered color labels for each timestep. The numbers indicate spatial information changes over time. Surprise effectively captures spatio-temporal properties, whereas alternative measure PMI does not perform as well. (b) presents the scenario with the information field with PMI values}
\label{fig:rolling_ball}
\end{figure*}

In this section, we provide a comprehensive step-by-step explanation of our technique's mechanism. We intend to construct an efficient, generic, and fast data summarization workflow with minimal customization to adapt to a variety of application domains. Figure \ref{fig:workflow} illustrates the schematic of this proposed workflow. To demonstrate each step of this workflow, we will refer to Figure \ref{fig:rolling_ball}, which serves as an illustrative application of our method using a synthetic data set. In this application, we simulate a rolling ball moving 0.5 units from left to right at each timestep until it exits the view area.

Our proposed method is designed for datasets with sequence of timesteps containing various types of salient spatial features. %. Each input \rmark{(input what? sample/point/?)}is referred to as a timestep. 
In Figure \ref{fig:rolling_ball}, the simulated rolling ball application consists of 19 timesteps (T0 - T18), showcasing the ball's positional change over time. Each timestep is represented as 800 x 400 pixel 2D RGB images and pixel intensities ranging from 0 to 255. At timestep T0, the frame is empty; the ball has not yet entered the view area. It enters at T1 and the ball changes position until T17. In, time step T18 the ball exits the view area. The method starts with iterating over input timesteps. After the first timestep, for every subsequent timesteps, the key regions are extracted using a  segmentation method proposed in~[\cite{7684170}]. Criteria for extraction of such key regions can be determined by the domain knowledge. In the rolling ball demonstration, the key feature is the presence of the ball and its location. So we segment the region containing the ball and create binary masked images shown in Figure \ref{fig:rolling_ball} (masked row). These images contain only two distinct data values: 0 (no ball) and 255 (ball).

After extracting the key region, we check if a certain property is present in that timestep. We denote this property as \textit{trigger} which is a change in the key region. The change can be in terms of count, size, shape, connectivity, space or association. %\rmark{Comment:give an example of a trigger here for clarity}.
In this rolling ball demonstration, the trigger is the presence of the ball in the view area. When the ball enters and exits the area, it is considered as a salient information. But the time it remains in the area, the only novel information is its change of position. If the trigger is present in the current timestep, then it is considered to be significant and a key timestep. Hence our method saves it as it is. If the trigger is not present in the current timestep, we proceed to the next timestep, do a similar check, and continue the process until a trigger event is encountered. These sequential time frames that did not have the trigger event are chosen to be fused into a single timestep as the amount of novel information within such a sequence is low. If the number of timesteps to be fused is one then we can discard it as the previous time step has the necessary information. If the number is greater than one then we perform pairwise information-guided fusion on these timestep and convert them as one single timestep to be saved as a temporal summary. Referencing the demonstration in Figure \ref{fig:rolling_ball}, T0 is saved. Then starting from timesteps T1 - T17 there is no trigger present in the timesteps. Therefore, those timesteps are fused as shown in figure \ref{fig:rolling_ball}(a) and (b). The following section sheds light on how we use information theoretic approaches for the fusion method.

  %Before diving into the motivation, we present the scenario where the trigger is not determined. In such a scenario, we can choose times for which we want to fuse the frames. In figure \ref{fig:paraview_rolling}(a), we have fused the first 5 time steps(T1-T5) and then 6 time steps(T6-T11) as an example. Again, these are arbitrary numbers for experimentation. The first row shows the data values (ranging from 0 to 255) highlighting the significant regions (ball in this case) based on the fusion conditions derived from pairwise maximum Specific Mutual Information SMI (explain in detail in the next 0section) values. We are referring to this image as Surprise Fused Information Field. In the next row, we present the  Surprise Fused Timestep Summary, where based on the max SMI values the time steps are labeled. The progression of the time is denoted in the color bar { \color{red} (Add the color bar)}. From this image, we can track how the ball is changing position and direction at every time step. 

%\subsection{Measuring Information Content of Data for Fusion Purpose}
\subsection{Characterization of Samplewise Information for Fusion}
In this work, our aim is to track and summarize the information change in fused timesteps. Therefore, we need to have a quantification of information content for each data point in the timesteps. We use the term "sample" to refer to these individual data points. Each timestep contains multiple samples representing the values of the data. For example, in the case of images like the rolling ball, these samples typically range from 0 to 255, while for other types of data variables, they may be scalar values. The quantification of information for these samples will help us identify important spatial features for the timestep. 

\subsubsection{Mutual Inforamtion}
In information theory, Mutual Information (MI) [\cite{Shannon1948}] is a prominent measure that estimates the total amount of shared information between two random variables. Given two random variables $X$ and $Y$, MI $I(X,Y)$ is formally defined as:
\begin{equation}
\operatorname{I}(X;Y)= \sum_{y\in Y} \sum_{x\in X} \operatorname{p}(x,y)\log \frac { \operatorname{p}(x,y)}{\operatorname{p(x)}\operatorname{p(y)}}\label{eq:MI}
\end{equation}
where $\operatorname{p}(x)$ and $\operatorname{p}(y)$ are the probabilities of occurrence of values $x$ for $X$ and $y$ for $Y$ respectively and $\operatorname{p}(x,y)$ is the joint probability of occurrence of values $x$ and $y$ together. MI assesses the degree of association or disassociation between two random variables and gives a single value. Since we aim to extract feature-based data summaries, we need samplewise spatial and temporal information characterization. Therefore, we leverage the decomposition of MI which quantifies each data value's contribution toward the association or dissociation. The decomposition of MI is termed as Specific Mutual Information or SMI [\cite{DeWeese1999}]. SMI measures the information content of the individual scalar values of one variable (reference) when another variable (target) is observed. This information quantification helps with identifying important spatio-temporal features in the data. There are multiple methods for MI decomposition [\cite{DeWeese1999,Butts2003}]. For apprehending the fusion criteria essential for summarizing the data, the properties of the SMI measure, \textit{Surprise}  holds the most potential.

\subsubsection{SMI Measure Surprise}
The Surprise measure denoted as $I_1$ was first introduced by \cite{DeWeese1999}. Surprise quantifies the information change of the target variable after observing the individual scalar values of the reference variable. The derivation of Surprise from MI is as follows. 

By definition the conditional probability of  $x$ given  $y$ or $p(x|y)$  is, 
\begin{equation}
\operatorname{p}(x|y) = \frac{ \operatorname{p}(x,y)}{\operatorname{p(y)}} \\
or \\  \operatorname{p}(x,y) = \operatorname{p}(x|y) \operatorname{p(y)}\label{eq:con}
\end{equation}

Replacing the joint probability in Equation \ref{eq:MI}, we get,
\begin{equation}
\begin{split}
\operatorname{I}(X;Y) & = \sum_{y\in Y} \operatorname{p(y)  \sum_{x\in X} \operatorname{p}(x|y)\log \frac { \operatorname{p}(x|y)}{\operatorname{p(x)}}} \\
& = \sum_{y\in Y} \operatorname{p(y)} \operatorname{I_1(y;X)} 
\end{split}
\label{eq:SMI_der}
\end{equation} 

where,

\begin{equation}
\operatorname{I_1}(y;X)= \sum_{x\in X}  \operatorname{p}(x|y)\log \frac { \operatorname{p}(x|y)}{\operatorname{p(x)}}\label{eq:surprise}
\end{equation}
Equation \ref{eq:surprise} represents the surprise measure of data value $y$ from $Y$  after observing all the values of $X$. A high value for $I_1(y;X)$ means after observing $y$, some previously low probable values of $x\in X$ have become highly probable in the distribution. This likelihood increase is the element of surprise and a salient finding for further analysis. Surprise is also the only positive decomposition of MI since it is the Kullback-Leibler distance between $\operatorname{p}(x|y)$ and $\operatorname{p(x)}$ [\cite{kullback1951information}]. In the fusion process of the data summarization, the data samples with high surprise values stand out and are identified as important features. 

\subsection{Surprise ($I_1$) Guided Fusion Technique}

%In the demonstration of the simulated rolling ball in Figure \ref{fig:rolling_ball}, timesteps T1 and T17 are fused. 
When the low informative timesteps are chosen,  the fusion initiates for summarization. The fusion is done on pairwise timsteps. For every pair of data samples from the timsteps, we store samples contributing to more information or high $I_1$ values. This fusion strategy was introduced in [\cite{Bramon2012}] for fusing different datasets to gain the most informative combination.  The condition to compute the fused value using ${I_1}$-fusion is:

For every data sample pair with $(x,y)$, the fused value, f is,

\begin{equation}
 \operatorname{f} = 
 \begin{cases}
    $x$,& \text{if } \operatorname{I_1(x;Y)} > \operatorname{I_1(y;X)}    \\
    $y$,              & \text{otherwise}
\end{cases}
\label{eq:fusion}
\end{equation}

Here $x \in X$ and  $y \in Y$ are individual data values from two data sets X and Y respectively. Our fusion criteria is based on the idea of Equation \ref{eq:fusion}, however, instead of different datasets we are using two subsequent timesteps from the same dataset. To fuse multiple timesteps, we begin by creating a fused timestep using the first two timesteps. Then, we repeat the fusion process by comparing the fused timestep with the next timestep, and continue until all desired timesteps have been fused. Our strategy involves updating the fused timestep during each iteration and selecting the spatial and temporal values with the highest information content. By the end of the process, the resulting fused timestep will represent a summary capturing their most informative properties with direction. The fusion process is described in detail in Algorithm \ref{algo:fusion}. After each fusion process, the algorithm provides 3 fused fields as shown in \ref{fig:rolling_ball}(a).  $I_1$ fused data value field contains the values of the data samples with high surprise measure. $I_1$ fused information value field contains the $I_1$ values for the same sample positions.  In the timestep summary fields, the same data sample are labeled with their originating timestep numbers. 

Applying the fusion process in Algorithm \ref{algo:fusion} on the timesteps T1 - T17 of the simulated rolling ball, the surprise $(I_1)$ fused data value field is generated highlighting the path of the rolling ball with data values 255 as shown in Figure \ref{fig:rolling_ball} a(i). The regions without the ball are valued 0 or white. Figure \ref{fig:rolling_ball} a(ii) represents the fused information fields with $I_1$ values. From the color bar's gradient, we observe that the surprise values exhibit limited variation, spanning approximately from 0 to 2. A minimal data value range results in minimal surprise value variation. Figure \ref{fig:rolling_ball} a(iii) presents the timestep summary where the numbers of originating timesteps are labeled for the salient samples. Here, we employed distinct colors to label the timesteps, enabling clear visualization and differentiation of each timestep. This color-coded representation effectively showcases the flow of information, facilitating the tracking of information changes over time. Here, the confidence threshold is employed to downplay the non-important regions. In this particular case, the threshold value is set to 255. Any value below 255, representing the absence of the ball, is assigned as timestep 0. In Figure \ref{fig:rolling_ball} a(iii), these regions are depicted as white or transparent (steps 19 - 24 in Algorithm \ref{algo:fusion}). The method reduces the number of output timesteps from 19 to 3 in the simulated rolling ball case, achieving substantial data reduction with minimal information loss. The fused timestep effectively visualizes the information changes over time, serving as a concise summary of the original data dynamics. 
\begin{algorithm}
\caption{Fusion Process}
\label{algo:fusion}
\hspace*{\algorithmicindent} \textbf{Input:} \vspace{-5px}
\begin{itemize}
    \item data1: Array of data values from fused timestep. Initialized with the first timestep. \vspace{-10px}  
    \item data2: Array of data values from the subsequent timestep.\vspace{-10px}
    \item Ifield1: Array of $I_1$ values for $I_1(x;Y)$ $\forall x \in X$ \vspace{-10px}
   \item Ifield2: Array of $I_1$ values for $I_1(y;X)$ $\forall y \in Y$\vspace{-10px}
   \item timestep\_fuse: Array of timestep values. Starts with 0\vspace{-10px}
   \item time: Current timestep value \vspace{-10px}
    \item conf\_th: Confidence threshold for the key regions. 
\end{itemize}
\hspace*{\algorithmicindent} \textbf{Output:} \vspace{-5px}
\begin{itemize}
    \item fused\_field\_data: Array of the fused data values \vspace{-10px} 
    \item fused\_field\_I1: Array of the fused $I_1$ values \vspace{-10px}
    \item timestep\_fuse: Array of the fused timestep values. 
\end{itemize}
\begin{algorithmic}[1]
\input{}
\Procedure{createFusionFields}
{\textit{data1}, \textit{data2}, \textit{Ifield1},
\textit{Ifield2}, \textit{timestep\_fuse}, \textit{time}, \textit{conf\_th}}
    \State $\textit{fused\_field\_data} \gets$ array of zeros with shape $\textit{data1}$
    \State $\textit{fused\_field\_I1} \gets$ array of zeros with shape  $\textit{data1}$
    \For{$i \gets 0$ to $\textit{data1.shape}[0] - 1$}
        \For{$j \gets 0$ to $\textit{data1.shape}[1] - 1$}
            \If{$\textit{Ifield1}[i][j] > \textit{Ifield2}[i][j]$}
                \State $\textit{fused\_field\_data}[i][j] \gets \textit{data1}[i][j]$
                \State $\textit{fused\_field\_I1}[i][j] \gets \textit{Ifield1}[i][j]$
                \If{$\textit{time} = 1$}
                    \State $\textit{timestep\_fuse}[i][j] \gets \textit{time}$
                \EndIf
            \Else
                \State $\textit{fused\_field\_data}[i][j] \gets \textit{data2}[i][j]$
                \State $\textit{fused\_field\_I1}[i][j] \gets \textit{Ifield2}[i][j]$
                \State $\textit{timestep\_fuse}[i][j] \gets \textit{time} + 1$
            \EndIf
        \EndFor
    \EndFor
    \For{$i \gets 0$ to $\textit{data1.shape}[0] - 1$}
        \For{$j \gets 0$ to $\textit{data1.shape}[1] - 1$}
            \If{$\textit{fused\_field\_data}[i][j] </> \textit{conf\_th}$}
                \State $\textit{timestep\_fuse}[i][j] \gets 0$
            \EndIf
        \EndFor
    \EndFor
    \State \textbf{return} $\textit{fused\_field\_data}, \textit{fused\_field\_I1}, \textit{timestep\_fuse}$
\EndProcedure
\end{algorithmic}
\end{algorithm}

\subsection{Alternative Fusion Approaches}
The measure Surprise effectively apprehends the spatio-temporal features for the data summarization process. However, we also explore several other potential information measures to devise alternative techniques for generating data summaries. These information-theoretic measures include Pointwise Mutual Information (PMI) [\cite{church1990word}], SMI measure (1) predictability $(I_2)$ [\cite{DeWeese1999}] and  (2) \textit{stimulus} specific information $(I_3)$ [\cite{Butts2003}]. These measures were investigated because they are the decomposition of MI and they possess the ability to analyze the contributions of individual data values in quantifying the information content of spatio-temporal data. They offer insights into the properties of data samples concerning information analysis.

\subsubsection{PMI Guided Fusion}

Pointwise Mutual Information (PMI) [\cite{church1990word}]  quantifies the degree of association (or disassociation) between individual data points given two variables. If $X$ and $Y$ are two variables, then each data point can be represented by the value pair $(x,y)$ where $x \in X$ and $y \in Y$. The statistical association between these two points can be measured by their PMI value, which is formally expressed as:

\begin{equation}
\operatorname{PMI}(x,y)= \log \frac { \operatorname{p}(x,y)}{\operatorname{p(x)}\operatorname{p(y)}}\label{eq:PMI}
\end{equation}

where $\operatorname{p}(x)$ and $\operatorname{p}(y)$ are the probabilities of occurrence of values $x \in X$ and $y \in Y$ respectively, and $\operatorname{p}(x,y)$ is the joint probability of occurrence of values $x$ and $y$ together. PMI was introduced by \cite{church1990word}. Comparing Equations \ref{eq:MI} and \ref{eq:PMI}, we can infer that the expected PMI values over all occurrences of variables $X$ and $Y$ correspond to the mutual information value $I(X;Y)$. PMI is a symmetric measure that can generate values ranging from negative to positive, depending on whether the distributions are complementary or overlapping. If the information overlap is high ($\operatorname{p}(x,y) > \operatorname{p}(x) \operatorname{p}(y)$ ), then $\operatorname{PMI}(x,y) > 0$.  The low  association is indicated by  $\operatorname{p}(x,y) < \operatorname{p}(x)\operatorname{p}(y)$, resulting in  $\operatorname{PMI}(x,y) < 0$. If $x$ and $y$ are statistically independent then $\operatorname{p}(x,y) = \operatorname{p}(x)\operatorname{p}(y)$ and  $\operatorname{PMI}(x,y) = 0$. 

Now, given the PMI measure, we can devise a fusion strategy similar to $I_1$ where $I_1$ values are substituted with PMI values in Algorithm \ref{algo:fusion}. The resulting fused information field on the simulated rolling ball is shown in Figure \ref{fig:rolling_ball}(b). We observe that PMI fails to capture the spatial characteristics of the key regions and only captures the overlapped regions that are strongly associative, indicated by high positive PMI values. The non-overlapping regions are transparent showing low information overlap. The PMI values here are 0 meaning data distribution is complementary and statistically independent. As the spatial position of sample pairs plays a critical role in PMI calculation, the fused field properties can exhibit significant variation depending on the degree of overlap between the key features. %This limitation is more apparent if they are further apart.%In the scenario of the simulated rolling ball, if the frames were sparser and exhibited less overlap in the ball regions between successive timesteps, the PMI measure would predominantly highlight the last timestep as a salient feature. The reduced overlap would result in reduced significance for previous timesteps in terms of PMI values. \cc{The detailed results are included in the supplementary section (if we decide to include the results).} 

\subsubsection{$I_2$ Guided Fusion} 
Predictability $(I_2)$ is another decomposition of MI introduced by \cite{DeWeese1999}. This SMI measure  quantifies the change in the uncertainty of one variable $(X)$ after observing the individual value of another variable ($y \in Y$) and is computed as:  
\begin{equation}
\operatorname{I_2}(y;X)= - \sum_{x\in X}  p(x)\log p(x) + \sum_{x\in X} p (x|y)\log p(x|y)\label{eq:predictability}
\end{equation}

where $y \in Y$ is the reference variable and $x \in X$ is the target variable. $\operatorname{p}(x)$ is the probabilities of occurrence of values $x$ for $X$ and $\operatorname{p}(x|y)$ is the conditional probabilities values $x$  given values $y$. Upon observing the variable y, the uncertainty of variable X can either increase or decrease, leading to the possibility of both positive and negative values for the $I_2$ measure. In some cases, the increased uncertainty can reveal significant information about the relationship between the variables. However, when we use the $I_2$ measure instead of the $I_1$ measure in the fusion process, the resulting fused information does not offer a meaningful summary over time. This measure intermittently highlights notable information changes, failing to capture the overall data dynamics. Our hypothesis is that this measure is more suitable for feature extraction and uncertainty quantification across various datasets rather than for analyzing consecutive timesteps within a single dataset.

\subsubsection{$I_3$ Guided Fusion} \label{I3_sec}

There is another measure that is derived from the decomposition of MI and was introduced for measuring the association of stimulus and response in certain neural systems [\cite{Butts2003}]. It is termed as Stimulus Specific Information (SSI), denoted by $I_3$ and computed as: 

\begin{equation}
\operatorname{I_3}(y;X)= - \sum_{x\in X} p(x|y) \operatorname{I_2}(x;Y)  
\label{eq:ssi}
\end{equation}

\begin{figure*}
    \centering
    \includegraphics[scale=.36]{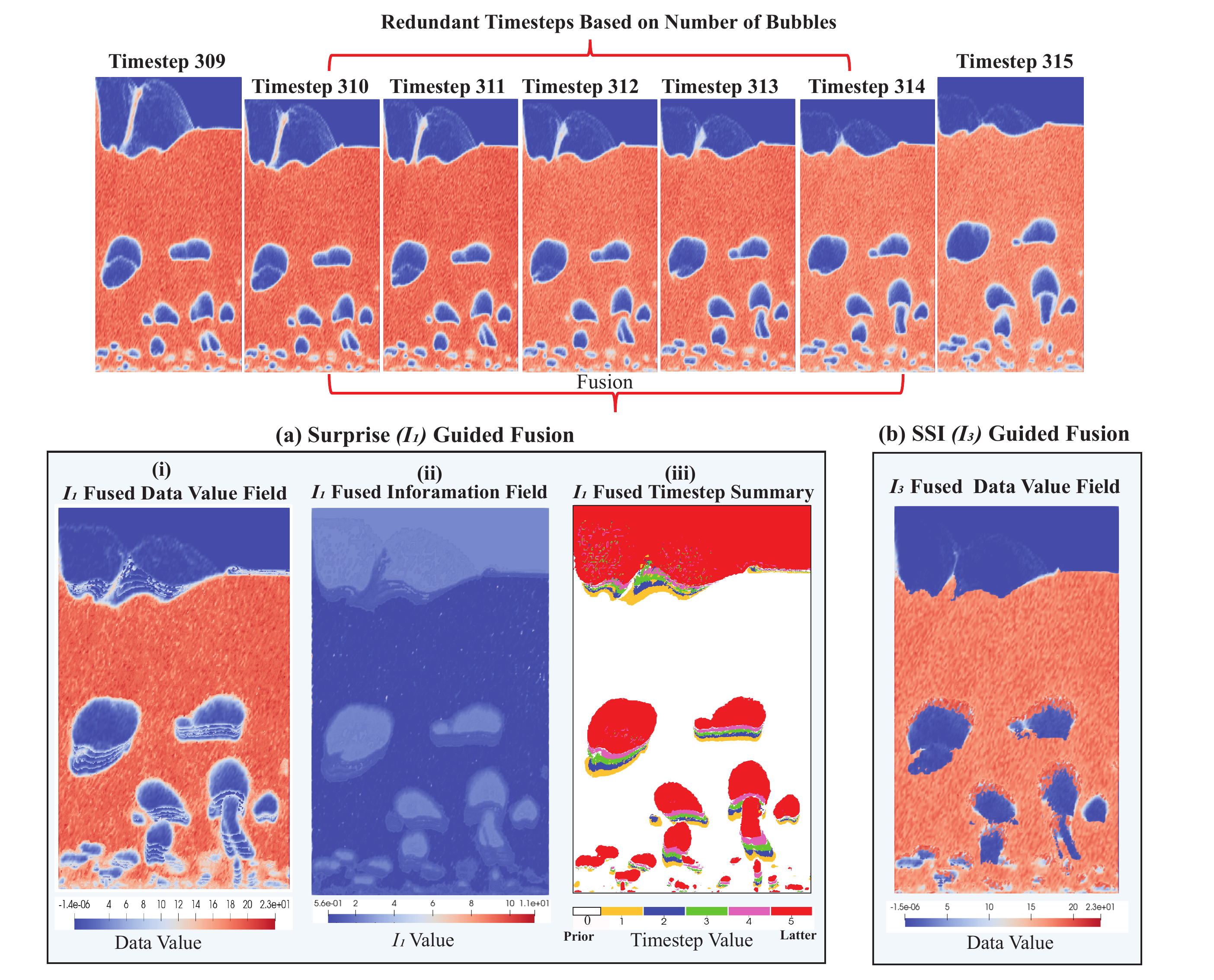}
    \caption{Analysis and illustration of dynamic spatio-temporal summarization for MFIX-Exa simulation application. The first row represents multiple timesteps (309 - 315) of the dataset where the bubbles are highlighted as the blue regions. Timesteps are raw images with 488 x 842 dimensions. Timesteps 310 to 314 are fused using (a) Surprise ($I_1$) guided fusion (i-iii) and alternative (b) Stimulus Specific Information (SSI) or $I_3$  guided fusion. a(i) shows the $I_1$ fused data value field of the timesteps, a(ii) shows $I_1$ fused information value field reflecting $I_1$ values and (c) shows  $I_1$ 5 fused timestep summary. The color bar labels 5 different colors summarizing the flow and direction of the bubbles in one timestep. The alternative $I_3$ is unable to capture the path of bubbles as reflected in (b) $I_3$ fused data field.}
    \label{fig:mfix}
\end{figure*}

The response and stimulus are the two variables X and Y. This measure focuses on establishing the association between the two variables to extract the maximum amount of information from their relationship. It emphasizes that the most informative data values from the first variable are related to the most informative data values of the second variable [\cite{Butts2003}]. %Since the $I_2$ measure can provide a reduction in uncertainty, averaging over the values with the most uncertainty reduction gives the most information. 
In some cases, $I1$ can be an alternate measure for $I3$, but the interpretation is different based on the data [\cite{Butts2003}]. They both capture different properties in the data. When we use the $I_3$ measure instead of the $I_1$ measure on the simulated rolling ball dataset, it captured very similar properties shown in Figure \ref{fig:rolling_ball}(a). However, when applied to a different and more complex dataset, it failed to capture the spatial properties of the features in the summarization. This is explained in detail in Section \ref{mfix_I3} and shown in Figure \ref{fig:mfix}(b). 

From the various methods mentioned, it's clear that different approaches emphasize distinct data properties. After thorough investigation of various information-theoretic methods, we found that the Surprise measure aligns best with our objectives of highlighting features and summarizing spatio-temporal data while tracking information flow. In the next section, we apply this method to more complex datasets to demonstrate its versatility across different data scenarios.  \label{method}
\section{Applications} 
In this section, we assess the versatility of our DSTS method across various multidisciplinary applications, demonstrating its effectiveness in handling complex and diverse datasets with streaming characteristics.

%By presenting these applications, we seek to highlight the broad applicability and robustness of the fusion method in capturing relevant information and tracking changes in different domains. 

We selected diverse applications across multiple domains to demonstrate the method's robustness. The first dataset involves multiphase flow simulation with scalar data values. The second is a surveillance video with extended timesteps to highlight optimization capabilities. The third dataset comprises complex immune cell images. This comprehensive evaluation showcases the method's adaptability and utility across various research disciplines.

%Through this examination, we aim to establish the method as a valuable tool for summarizing and analyzing spatiotemporal data, offering researchers a versatile approach for understanding information flow and extracting key features from their datasets. 

\subsection{MFIX-Exa Flow Simulation} \label{sec:mfix}
MFIX-Exa [\cite{Musser2022}] is a particle-based multiphase flow simulation developed by the National Energy Technology Laboratory (NETL), USA. It focuses on modeling multiphase flow by interacting with a vast number of particles within the simulation domain. Using MFIX-Exa, particle-based data is generated for studying the operational principles of chemical looping reactors (CLR). MFIX-Exa can simulate various chemical mixing processes and one such simulation involves the interaction of carbon particles with air in a fluidized bed, resulting in the formation of carbon dioxide bubbles or void regions. These bubbles hold significant importance for domain experts.% conducting these simulations.
MFIX-Exa generates vast data with numerous particles and thousands of timesteps. This extensive raw data presents challenges in transferring to storage due to limited I/O bandwidth. Hence, experts seek solutions to extract bubble-specific information while reducing storage needs, and preserving temporal bubble dynamics. Our proposed dynamic spatio-temporal summarization method can be used to provide this solution. %to fulfill this requirement where a highly compressed data summary will be obtained while preserving the temporal bubble dynamics.  

%(\rmark{This following paragraph needs to be re-written.. it should be made clear that we are working on an image-based data, the processing of the original particle data to generation of images are not done in this paper, but you can briefly mention the process and refer the details of density field generation to this paper: Feature Analysis, Tracking, and Data Reduction: An Application to Multiphase Reactor Simulation MFiX-Exa for In-Situ Use Case. Currently it reads like we are doing all the processing as part of our method...})
\subsubsection{Data Context and Features}
The raw data of the MFIX-Exa simulation contains the value of particle properties (i.e., position, velocity, momentum, etc.) over multiple time steps. As the data is streaming in nature, therefore it is a good fit for our proposed workflow. In \cite{Biswas2021}, there is a detailed description of how this streaming simulation is converted into a 3D scaler dataset. % The key regions We start by extracting the key regions which are the \textbf{"bubbles"} in this case. We segment the feature: bubbles. 
%The data is converted into a 3D scalar particle density field. This transformation involves creating a 3D histogram using the particle locations as inputs. The bin frequencies within the histogram indicate the particle count in specific local regions of the domain. Subsequently, this 3D spatial histogram is converted into a structured grid dataset, where the dimensions of the grid align with the spatial dimensions of the histogram bins. The frequency values within each grid point then represent the particle density at that particular location. 

For this work, we use 2D slices extracted from this raw 3D scalar data. These slices contain scaler values representing particle density. As a result, our dataset consists of multiple timesteps (count 332), and each timestep corresponds to 2D data samples with dimensions of 488 x 842. The sample values fall within the range of [-\num{1.2e-6}, 29.08]. As mentioned earlier, the key regions are the \textbf{bubbles}. The bubbles are the low particle density regions. In \cite{Dutta2022}, the detection, segmentation, and characterization of the bubbles are developed and studied in an extensive manner. Using a density threshold, the bubbles are segmented. Then, the VTK [\cite{schroeder1998visualization}] library is used to extract the connected components in the dataset and use the threshold value to filter the low-density bubbles. Over time, the bubbles undergo phases like creation, merge, split, and dissolve into air %(indicated by going out of the simulation domain)
[\cite{Biswas2021}]. Using data from this simulation, domain experts want to comprehend the evolution of bubbles and explore the relationships between various bubble characteristics such as their size, shape, number of bubbles, etc [\cite{Dutta2022}]. Important events in this simulation can be characterized by the creation of a bubble, the merging of two or more bubbles, or the dissolving of a bubble. Note that for all of these events, the total number of bubbles will change. Hence, if we identify time steps when the bubble number has changed from a previous time step, then that can be considered as a "trigger". In this work, we skip counting changes in very small bubbles for simplicity. Key time steps are saved when the trigger occurs, and intermediate steps between two triggers are fused using Algorithm \ref{algo:fusion} to summarize and store them. This process continues for all timesteps.

\subsubsection{Results for Data Summarization} \label{mfix_I3}
%one sentence summary the surprise method (figures) is superior to the I3 method. Surprise represents the information flow visually better..
Figure \ref{fig:mfix} shows the analysis of DSTS method for MFIX-Exa simulation. Timesteps 309 - 315 are shown in the first row where bubbles are the blue regions. From 311 to 314 the number of bubbles remains unchanged. Therefore timesteps 310 - 314 are summarized using the fusion process. Figures \ref{fig:mfix} (a) represent results from the Surprise $I_1$ guided fusion. The  $I_1$ fused data value field a(i) shows the scaler values [\num{-1.4e-6} to 23] for the fused timesteps. Here the change in bubble movement is very prominent. Figure \ref{fig:mfix} a(ii) presents the $I_1$ values [\num{5.6e-1} to 11] for the fused timesteps. The range of information value ($I_1$) is smaller, making it less sensitive to bubble movement compared to particle density values. However, it effectively highlights the main bubbles and their temporal dynamics. The timestep summary field depicted in Figure \ref{fig:mfix} a(iii) represents the timestep values from which the bubbles originate. Here 5 timesteps are color-coded in distinguished colors to highlight the flow of the information between interacting bubbles. The color map reflects the direction of the bubbles from the start to the end position. This summary field not only emphasizes key features but also visually indicates the spatial information flow over time. The white/ transparent background (labeled as 0) filters all the density values that are of low importance for this dataset. %In Algorithm \ref{algo:fusion}, the thresholding is implemented in lines 19 - 25 by specifying the threshold value "conf\_th". Since bubbles are of low density, any values > "conf\_th" are assigned as 0 to tone down the non-bubble regions. Those regions are white in the timestep summary fields and denoted as timestep 0.

%mention specifically that it is an alternative method. make more negative. 
We also implemented the alternative SSI ($I_3$) guided fusion technique as mentioned in Section \ref{I3_sec} for MFIX-Exa. Figure \ref{fig:mfix}(b), represents the $I_3$ fused data value field. Here the bubbles are somewhat highlighted but the change in the bubbles' movement is hard to interpret. $I_3$ captures some spatial features of the bubble regions but the edges are blurred. Thus, the surprise fusion method again proves to be better than the SSI measure.

Based on the outcomes observed in both the simulated rolling ball and MFIX-Exa simulation applications, it is evident that the timestep summary field stands out as the most informative visual representation for spatio-temporal summarization.  This representation encapsulates both spatial and temporal dynamics by highlighting the key regions and indicating the direction of the information flow over time. Consequently, for our analysis of the next two applications, we only show the surprise fused \textbf{timestep summary fields}.
%\iffalse
%This code uses the Visualization Toolkit (VTK) library to perform segmentation on a dataset.

%The code creates two objects of the vtkConnectivityFilter class named seg and segmentation.

%The first vtkConnectivityFilter object, seg, is used to extract the largest connected region in the dataset. It takes the output of a threshold filter (represented by the thresholding variable) as input, and sets the extraction mode to the largest region. Then, it updates the filter to perform the segmentation and store the result.

%The second vtkConnectivityFilter object, segmentation, is used to extract all connected regions in the dataset. It also takes the output of the thresholding filter as input, sets the extraction mode to all regions, turns on region coloring, and updates the filter to perform the segmentation and store the result.

%Overall, this code is segmenting a dataset by extracting connected regions based on a threshold and storing the results in two different vtkConnectivityFilter objects.

%\fi

\subsection{Surveillance Data Analysis and Optimization } \label{sec:SBM}
Data summarization techniques are vital for security camera footage analysis. Security camera systems generate vast amounts of data, and manually reviewing the continuous stream of video can be time-consuming and laborious. DSTS offers an efficient solution by allowing the analysis and optimization of camera footage. Suspicious activities can be detected by choosing an appropriate "trigger" and the generated summary fields provide experts with an intuitive visual representation of the key timesteps. The most significant impact of our method in this application is on archiving and storage optimization.

To demonstrate DSTS in security camera footage analysis, we used the publicly available SBM-RGBD Dataset [\cite{SBM-RGBD, Camplani2017}]. This dataset was originally created in conjunction with the Workshop on Background Learning for Detection and Tracking from RGBD Videos [\cite{back_detec}]. It was created to evaluate background modeling methods for moving object detection. The dataset comprises 33 RGBD videos, totaling 15033 timesteps, recorded indoors using a Microsoft Kinect sensor [\cite{Camplani2017}]. The dataset contains videos capturing moving objects at intervals, which aligns with the data requirements for our application. Here, we used one of the videos titled "Multipeople2" which shows four individuals walking in and out of the view area, engaging in discussions, and writing on a whiteboard. % The movements of four people are recorded as they enter and exit the meeting room. 
Our method shows an effective demonstration of summarizing the movement patterns of the individuals.

%The SBM-RGBD Dataset used in this application differs from our MFix dataset discussed in Section \ref{sec:mfix}. The SBM-RGBD Dataset comprises RGBD videos captured by a camera, representing image space data. On the other hand, the MFix-Exa simulation involves particle density values, representing scalar data. In the following section, we describe the effectiveness of the method in an image space. 
\begin{figure*}
	\centering
		\includegraphics[scale=.40]{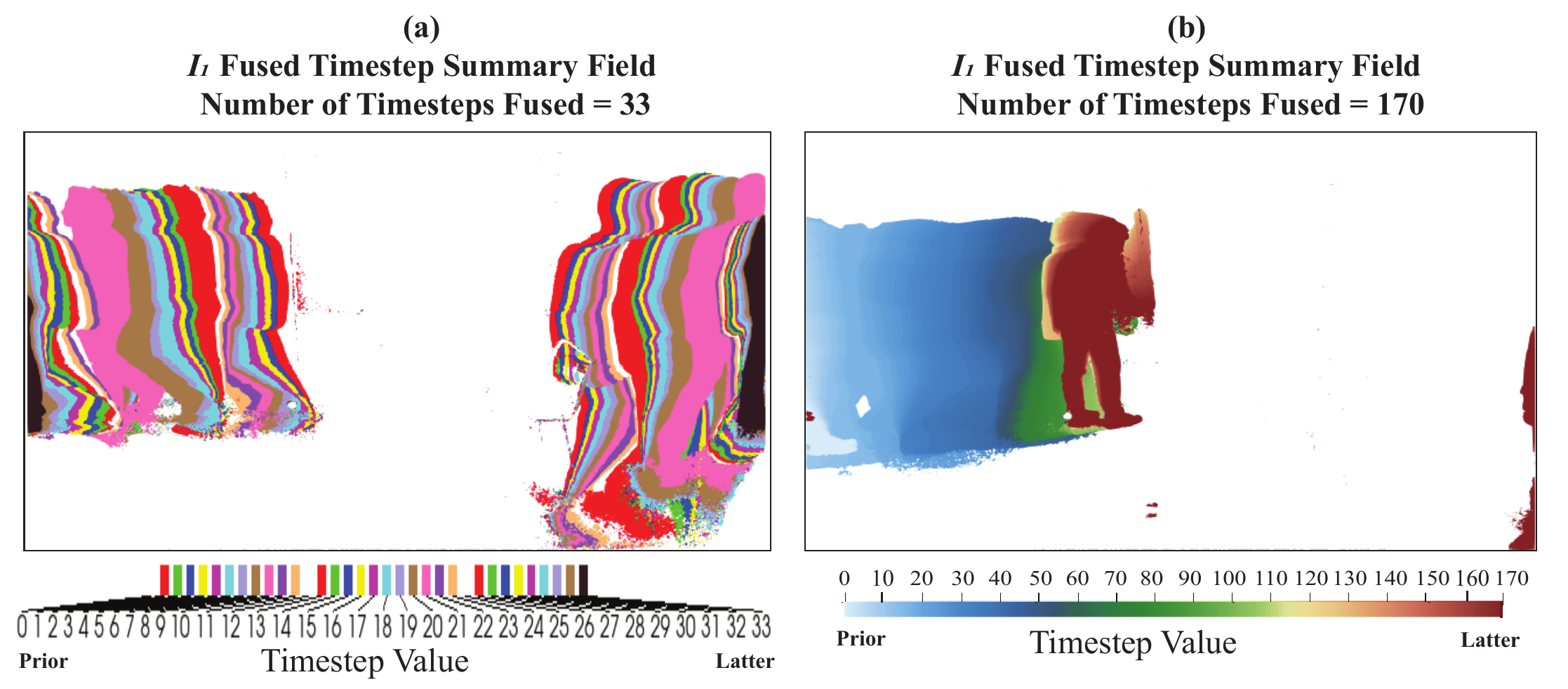}
	\caption{Analysis and illustration of the dynamic spatio-temporal summarization for the SBM-RGBD dataset. Here the emphasis is on representing the extensive number of timestep summaries.  (a) shows a summarization of 33 fused timesteps using a discrete color bar to illustrate information changes over time. (b) showcases a summarization of 170 fused timesteps, employing a continuous color bar to depict information changes over a longer period. The color bars point out the spatial direction of the information flow by denoting the prior and latter states.}
	\label{fig:multipeople}
\end{figure*}
\subsubsection{Data Context and features}
The videos have 640 x 480 resolution and the length is 1400 timesteps. The dataset comes with image (PNG) files for each timestep of the input video. We recreated the videos from the images which can be visualized here[BC lab ref]. The key regions in this dataset are the individuals and their movement. The whiteboard and a chair are stationary in the background. To extract key regions from the background, we have used a method called ViBe: a universal background subtraction algorithm designed for video sequences [\cite{barnich2010vibe}]. The algorithm aims to identify moving objects within consecutive images or videos by efficiently distinguishing between the foreground (moving objects) and the background (stationary elements). The algorithm is computationally lightweight, making it suitable for real-time applications.% It maintains a background model by continually updating pixel values based on temporal and spatial information. ViBe uses random sampling and background pixel voting to classify pixels as background or foreground. By comparing the current pixel value with its corresponding background model, ViBe determines if a pixel belongs to the foreground or the background. Moving objects results in a deviation from the background model, which allows ViBe to detect and segment them accurately.
The adaptability of ViBe has made it a popular choice for a range of computer vision tasks, including object tracking, surveillance, and motion detection. Its straightforward pseudocode provided in \cite{barnich2010vibe} facilitates easy implementation. By applying the ViBe algorithm, the RGB images are converted into masked binary images with segmented individuals in the foreground. 

In scenarios with multiple individuals in the frame, we adopt a concept similar to that used for counting bubbles in the MFIX-Exa (Section \ref{sec:mfix}). We apply the concept to count the number of people in each timestep by analyzing the largest connected regions in the masked images.%, utilizing the VTK library [\cite{schroeder1998visualization}]. 
Since it is a binary image, the data samples have two values: 0 (no individual) and 255 (individual). By setting a size threshold, we can accurately count the number of individuals in each timestep. Given that individuals move in and out of the view area in the dataset, the number of individuals can be used as a trigger for our application. Whenever a person enters or exits, that is a key timestep. The consecutive timesteps between two triggers are then fused using Algorithm \ref{algo:fusion}. This fusion process effectively summarizes the movement patterns of individuals within one timestep, eliminating the need to store every less informative timestep.

The combination of the key and summary timesteps, provides an intuitive visual representation of the significant moments in the video, making the analysis of surveillance scenarios more informative.   

\subsubsection{Results for Data Summarization}
Figure \ref{fig:multipeople} showcases the summarization fields for two separate fused timesteps in the dataset. Both fields show Surprise($I_1$) fused timestep summaries, effectively representing spatial features and movement directions of individuals over time.In this particular application, our main aim is to showcase how effectively the method can fuse longer timesteps while ensuring that both spatial and temporal dynamics remain just as noticeable as they are in shorter timesteps.

In Figure \ref{fig:multipeople}(a), the summarization field depicts a fusion of 33 timesteps, where two individuals are walking out of the view area. The leftmost person exits first, initiating the trigger and stopping the fusion process. Each timestep is represented by a discrete color, highlighting the changes in movement over the 33 timesteps. Sample values below 255 are set to 0, representing the white background, as the data value of individuals is 255.

Figure \ref{fig:multipeople}(b) shows a timestep summary for a longer period of 170 timesteps. The summarization field captures an individual walking into the view area and writing on the board, while another person has just stepped in, initiating the trigger event. Continuous colors are used to display the movement changes due to the length of the fused timesteps. 

Remarkably, key and summarized timesteps in this dataset result in a significant reduction from 1400 timesteps to only 49 timesteps. The highest number of timesteps being fused is 262. This notable optimization ensures that only relevant information is stored, reducing storage requirements without compromising critical insights into the movement patterns of individuals.  By efficiently identifying and storing key moments, our method enhances the analysis of crowded environments, enabling rapid detection of suspicious activities, and thereby serving as a valuable tool for surveillance.

\begin{figure*}
	\centering
		\includegraphics[scale=.35]{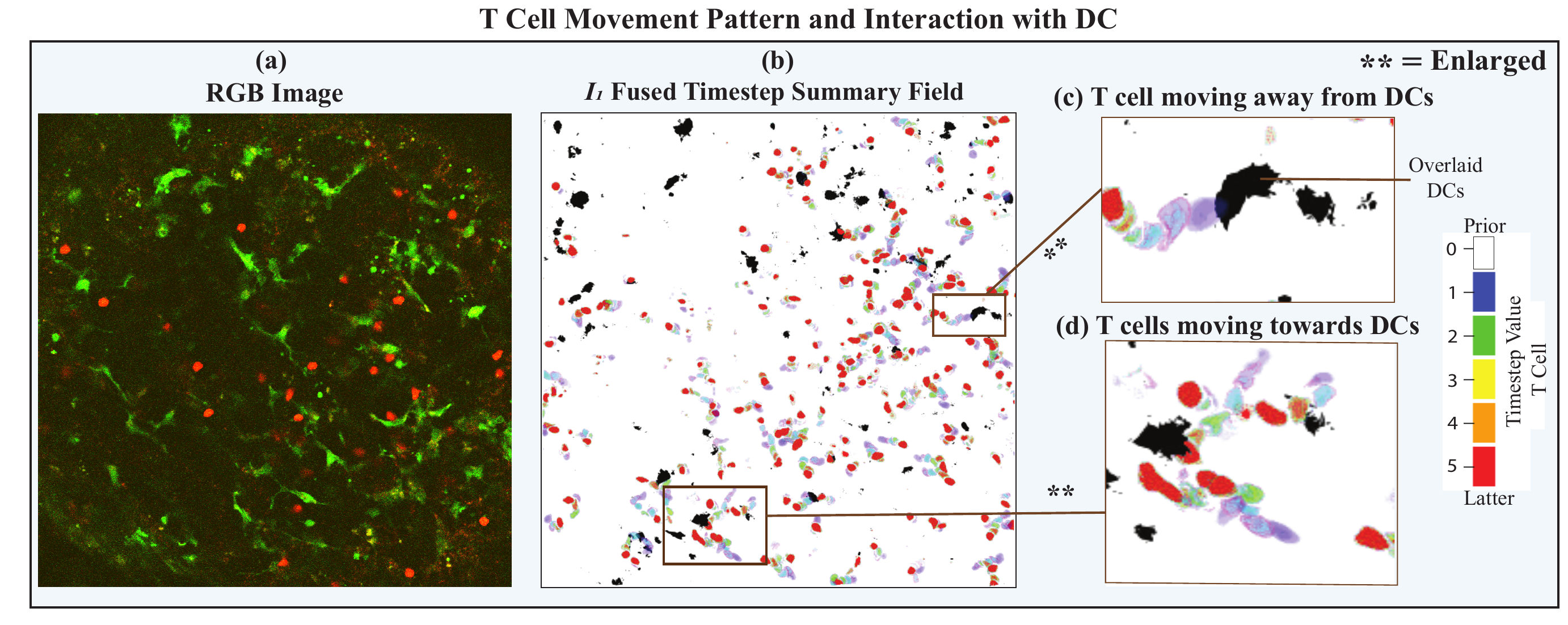}
	\caption{Illustration and analysis of the dynamic spatio-temporal summarization method for cell interaction in Lymph Node. The results emphasize T cell movement patterns and interaction with DCs in LN. (a) is a sample timestep. Here red indicates the T cells and green indicates the DCs. (b) is the surprise fused timestep summary fields of T cell for 5 timesteps. The black cells in the field are overlaid DCs to highlight cell contact. (c) and (d) are 2 enlarged positions from the (b) field to emphasize T cell movement. (c) shows that a T cell is moving away from the DCs. (d) shows multiple T cells moving toward the DCs. The corresponding timestep values are provided in the color bars to highlight the direction.} %In (ii), image (c) is a sample timestep for another T:DC dataset. (d) shows  summarization field for 6 fused timesteps from that dataset. Here also the DCs  (black) are overlaid. The enlarged images $d_1$ and $d2$ point out two scenarios focusing on multiple and minimal cell interaction respectively. These interactions are critical to the trigger detection process for this dataset.  } 
	\label{fig:cell}
\end{figure*}

\subsection{Tracking Cell Interactions in Lymph Nodes}
This dataset contains consecutive images of cellular interactions within the lymph node (LN). LNs are essential for immune function, playing a crucial role in initiating immune response and facilitating immune cell communication [\cite{mirsky2011systems}]. In the LN micro-environment, naive T cells are activated by interactions with different cell types. Understanding these interactions provides valuable insights into immune activation [\cite{brewitz2017cd8+}]. We reanalyze data from [\cite{Tasnim2018}] where information theory-based approaches were used to identify and quantify the spatial relationships between naïve T cells and three target cellular components: dendritic cells (DCs), fibroblastic reticular cells (FRCs), and blood vessels (BVs). These interactions are critical in influencing T cell movement and the timing of encounters with antigen-presenting DCs. This process is a key step in T cell activation and the initiation of the adaptive immune response.

The data for the study was gathered using two-photon microscopy (2PM) [\cite{rubart2004two}] to acquire 3D image stacks of lymph node tissue samples from mice. The imaging process captured dynamic movies lasting 10 to 45 minutes, resulting in a sequence of 3D images over time. This dataset is well-suited for the application of our method.

In the following sections, we demonstrate that our approach  offers both quantitative analysis and visualization of cell movement and communication. Additionally, this dataset showcases the extension of our technique's applicability to 3D images and movies, highlighting its versatility and suitability for complex spatial interactions in biological systems.

\subsubsection{Data Context and Features}
%Before performing the two-photon microscopy imaging the cells are dyed for identification in the tissue sample. Therefore, as shown in 

Figure \ref{fig:cell}(a) shows  2 RGB 3D images with  T cells dyed red and Dendtritic Cells dyed green.  Each voxel contains the color intensities of the dye in the red, blue, and green channels. For every time step, we extract the red and green channels into two separate images. We focus on the red channel in order analyze T-cell motility.

%, we will implement our technique on the T-cell which is the red channel in the T:DC interaction dataset shown in Figure \ref{fig:cell} (a) and Figure \ref{fig:cell} (c). 

%Figure \ref{fig:cell}(a) is a timestep of T:DC dataset with  512 x 512 x 22 dimensions. This dataset has 51 timesteps. Figure \ref{fig:cell} (c) is also a T:DC dataset with 1024 x 1024 x 23 dimension size and 32 timesteps.      

%The process of dyeing the cells and using two-photon microscopy leads to 

%The images contain significant noise, making this dataset more complex than the ones discussed in sections \ref{sec:mfix} and \ref{sec:SBM}. To address this issue, 

Because these images contain a lot of noise, we implemented a pre-processing step using the median filter [\cite{gonzales1987digital}], to reduce noise while preserving the edges of the cells for improved visualization. Since the red channels specifically represent the T cells we did not need to segment the data. 
%These cells are the key regions of interest for our analysis, and by focusing on the red channel, we can efficiently study the dynamics and movement patterns of the T-cells without the need for further segmentation. 

Our goal is to visualize how T cells move and interact in these movies. %Visualizing and quantifying these interactions is crucial and of interest to experts for further analysis. 
%The data source paper
In [\cite{Tasnim2018}] we used mutual information (MI) and normalized mutual information (NMI) to quantify associations between cells. %In this work, we are not comparing the association of cells across different mice or LN positions. Therefore, we are using
Here for each timestep, we use the MI value between two cell types as a "trigger" for fusing time steps. If the MI value for a specific timestep exceeds a specified threshold, we save that as key timestep. If the MI value falls below the threshold, we find the next timestep in which the MI value exceeds the trigger threshold and fuse the intermediate ones. %A lower association is indicated between cells for these fused timesteps. 
This allows us to efficiently capture and represent significant interactions while fusing less informative time steps. This provides an informative summary of  %facilitating a more insightful analysis of 
T cell interactions within the dataset. %The details of the MI calculations for the trigger are discussed in the following section. 

%Each timestep in this dataset presents multiple interactions of cells. 

%%One challenge is that the interactions between T cells and DCs, are sporadic, and there is a lot of empty space (black regions) indicating no cells in the RGB images shown in Figure \ref{fig:cell}. 
 
 %%MI calculation for each timestep between 2 cell types gives an average value for the association. For our dataset, calculating the mutual information (MI) value for one timestep can be challenging as a trigger. Non-interacting and non-cell locations minimize the total MI value, and differences with the next timestep may be negligible. 
 
 %%NOTE: Can you explain why this is a challenge? Do you mean that in a region with few interactions, you want to capture even small MI? 

 %%To address this, we divided both red and green channels into regions based on the T-cell diameter (5 - 7 microns), using 6 x 6 micron regions or 5x5 pixels for simplicity.
 
 %%NOTE: I don't understand what the size of the regions is and how its related to the T cell size. 
 
 %%For each timestep, we calculated the MI between red and green cells for each region, and the maximum MI value across all regions was used as the MI for the corresponding timestep.

%%NOTE: I also don't understand the previous sentence.
 
 %%Further detailed discussion on the regionalization method can be found in \cite{Tasnim2018}. Figure \ref{fig:cell}(ii) provides a detailed illustration of this scenario which is discussed in the next section.

\subsubsection{Results of Data Summarization}

Figure \ref{fig:cell} focuses on the T cell movement pattern and interaction with DCs in the summarized timesteps. Figure \ref{fig:cell}(a) is a sample timestep of T:DC dataset with  512 x 512 x 22 dimensions. This dataset has total 51 timesteps.    %we can observe the interactions between T cells and antigen-presenting Dendritic cells. 
Figure \ref{fig:cell}(b) displays the Surprise ($I_1$) fused timestep summary field, representing five fused timesteps from this dataset. The black cells in the summarization represent the Dendritic cells' value field overlaid on the fused summary field, visually illustrating the physical interactions between T cells and antigen-presenting DCs. Given that there are multiple interactions captured in each timestep, we highlight two specific interactions by enlarging the locations in Figures \ref{fig:cell}(c) and (d). In Figure \ref{fig:cell}(c), we observe a T cell moving away from the DCs. The color bar on the right side indicates the first (blue) and last (red) timesteps in the summary field, clearly indicating the movement direction. In Figure \ref{fig:cell}(d), we see multiple T cells moving toward the Dendritic cells, with cells making explicit contact with the DCs. The visualization provided by the fused timestep summary field allows for a comprehensive understanding of the dynamic interactions between T cells and DCs, providing valuable insights into the temporal dynamics of immune cell communication.

Our method, %on the other hand, successfully quantifies and most importantly 
successfully visualizes physical contact between cells and track movement over time. Other studies [\cite{dunn2011practical,mrass2017rock, mempel2004t}] using similar datasets have presented the statistical quantification of association. This addition of the visualization feature has the potential to unveil more insights for experts to analyze these associations for further investigation. This is a notable contribution to the study of T cell motility, a crucial aspect for understanding immune response dynamics.

    \label{results}
\section{Discussion} 

The proposed DSTS technique has demonstrated its effectiveness and versatility across various applications. Starting with a simple simulation of a rolling ball to analyzing complex cellular interactions within lymph nodes, the method proved its robustness for every application scenario. The combination of the key and fused timestep resulting from the method provides an impactful data summarization. The intuitive visual representation is a plus in showcasing an ideal blend of data optimization while preserving vital information change over time. We investigated multiple information theory measures and established that SMI measure surprise has the ability to capture the spatio-temporal features effectively.

We selected applications from multiple domains to shed light on different aspects of the DSTS  method. This technique offers a practical solution to downsample and analyze the vast particle data (bubbles) generated by the MFiX-Exa simulation. This application analysis focuses on the fact that the method is able to handle scaler value data as well as image data that incorporates the rest of the applications.  %It efficiently identifies and tracks bubbles within the simulation, down-sampling the data while preserving essential features. 

The RGBD tracking dataset was introduced to show the method's ability to summarize and highlight important movement patterns of individuals for longer periods of time. The results from this dataset reflect that longer fused timesteps are equally apprehensible as the shorter timesteps. This has promising implications for surveillance and security applications, enabling more efficient and informative analysis of large video datasets.

%In the case of surveillance camera footage, the method's ability to summarize and highlight important movement patterns of individuals provides significant advantages over traditional post hoc analysis and visualization approaches. By using triggers based on the number of people in the frame and fusing timesteps between triggers, the approach successfully reduces storage requirements while preserving essential insights into human movement within the scene. This has promising implications for surveillance and security applications, enabling more efficient and informative analysis of large video datasets.

We increased the complexity of the dataset gradually to prove the method's scalability. The T cell and Dendritic cell interactions in lymph nodes is a complex dataset. T cells which are the key regions were ample in number and the interactions with DCs were sporadic in nature. Given the fact that there were multiple interactions, our method was able to track every one of them. The summarization highlighted immune cell communication by providing a clear and comprehensive visualization of the T cell movement.  The cell interaction visualization opens up new possibilities for immunological research. 

All the applications in this work present post hoc data analysis. The datasets were already available when the method was applied. Since the method is not computationally expensive it can be easily combined as a step for analyzing data when they are generated; which is termed as in situ analysis. Through the integration of this method in any situ analysis, the resulting data will be optimized in real-time ensuring storage reduction with minimal data loss.

 \label{discussion}
\section{Conclusion and Future Work}
While the method's performance on these datasets is very promising, challenges may arise in selecting appropriate triggers and threshold values, especially in complex datasets with multiple key features, interactions, and noise. However, the flexibility of the technique allows for the adjustment of parameters to tailor the summarization process to different applications. Additionally, future research could explore combining different information-theoretic measures to further enhance the summarization capabilities and address specific challenges in various datasets.

In conclusion, the proposed dynamic spatio-temporal summarization technique offers a powerful and efficient tool for visualizing and analyzing complex time-varying datasets across different domains. The method has demonstrated its adaptability and potential to provide valuable insights and understanding of temporal data dynamics. The approach holds promise for advancing research in various fields which may lead to novel discoveries and applications, ultimately contributing to a deeper understanding of complex data systems.\label{conclusion}

\printcredits

\section*{Declaration of Competing Interest}
The authors declare that they have no known competing financial interests or personal relationships that could have appeared to influence the work reported in this paper.

\section*{Data Availability}
Data will be available upon request to the corresponding author. 

\section*{Acknowledgements}
We thank Judy Cannon for providing the biological data; Matthew Fricke for help on the cell interaction and analysis; Moses Biocomputational Lab at UNM for helpful discussion; IIT Kanpur..

\bibliographystyle{cas-model2-names}

\bibliography{timestep_summary}
\iffalse


\bio{}
Author biography without author photo.
Author biography. Author biography. Author biography.
Author biography. Author biography. Author biography.
Author biography. Author biography. Author biography.
Author biography. Author biography. Author biography.
Author biography. Author biography. Author biography.
Author biography. Author biography. Author biography.
Author biography. Author biography. Author biography.
Author biography. Author biography. Author biography.
Author biography. Author biography. Author biography.
\endbio

\bio{figs/pic1}
Author biography with author photo.
Author biography. Author biography. Author biography.
Author biography. Author biography. Author biography.
Author biography. Author biography. Author biography.
Author biography. Author biography. Author biography.
Author biography. Author biography. Author biography.
Author biography. Author biography. Author biography.
Author biography. Author biography. Author biography.
Author biography. Author biography. Author biography.
Author biography. Author biography. Author biography.
\endbio

\bio{figs/pic1}
Author biography with author photo.
Author biography. Author biography. Author biography.
Author biography. Author biography. Author biography.
Author biography. Author biography. Author biography.
Author biography. Author biography. Author biography.
\endbio

\fi
\end{document}